# NeuralBoneReg: An instance-specific label-free point cloud-based method for multi-modal bone surface registration

Luohong Wu [a,*], Matthias Seibold [a], Nicola A. Cavalcanti [a], Yunke Ao [a,b], Roman Flepp [a], Aidana Massalimova [a], Lilian Calvet [a], Philipp Fürnstahl [a,b]

[a] Research in Orthopedic Computer Science, Balgrist University Hospital, University of Zurich, Lengghalde 5, Zurich, 8008, Zurich, Switzerland
[b] AI Center, ETH Zurich, Ramistrasse 101, Zurich, 8092, Zurich, Switzerland

ABSTRACT

**Background:** In computer- and robot-assisted orthopedic surgery (CAOS), patient-specific surgical plans are generated from preoperative medical imaging data to define target locations and implant trajectories. During surgery, these plans must be precisely transferred to the intraoperative setting to guide accurate execution. The accuracy and success of this transfer rely on cross-registration between preoperative and intraoperative data. However, the substantial heterogeneity across imaging modalities and devices renders this registration process challenging and error-prone, leading to inaccuracies. Consequently, more robust and accurate methods for automatic, modality-agnostic multimodal registration of bone surfaces would have a substantial clinical impact.

**Methods:** We propose NeuralBoneReg, an instance-specific self-supervised, surface-based framework for bone surface registration using 3D point clouds as an intermediate representation. NeuralBoneReg comprises two key components: an implicit neural unsigned distance field (UDF) module and a multilayer perceptron (MLP)-based registration module. The UDF module learns a neural representation of the preoperative bone model. The registration module solves both global initialization and local refinement by generating a set of transformation hypotheses to register the intraoperative point cloud with the preoperative neural UDF based on a coarse-to-fine strategy. Compared to state-of-the-art (SOTA) supervised registration, NeuralBoneReg operates in an instance-specific self-supervised manner, without requiring inter-subject training data with ground truth transformations. We evaluated NeuralBoneReg against baseline methods on two publicly available multi-modal datasets: a CT–ultrasound dataset of the fibula and tibia (UltraBones100k) and a CT-RGB-D dataset of spinal vertebrae (SpineDepth). The evaluation also includes a newly introduced CT–ultrasound dataset of cadaveric subjects containing femur and pelvis (UltraBonesHip), which will be made publicly available.

**Results:** Quantitative and qualitative results show that NeuralBoneReg achieves competitive performance across anatomies and modalities. On UltraBones100k, it obtains an RRE of $1.83 \pm 1.30°$, an RTE of $2.02 \pm 1.30$ mm, an RR of 0.89, a CD of $0.82 \pm 0.12$ mm, and an HD95 of $2.06 \pm 0.36$ mm. On UltraBonesHip, it maintains stable performance with an RRE of $1.90 \pm 1.56°$, an RTE of $2.21 \pm 0.86$ mm, an RR of 0.88, a CD of $2.50 \pm 1.08$ mm, and an HD95 of $8.97 \pm 4.08$ mm, while other methods degrade significantly. On SpineDepth, it achieves an RRE of $3.78 \pm 19.34°$, an RTE of $2.80 \pm 3.75$ mm, an RR of 0.84, a CD of $1.78 \pm 1.61$ mm, and an HD95 of $4.26 \pm 4.58$ mm. Overall, the method consistently achieves accuracy close to pseudo ground truth across datasets.

**Conclusion:** NeuralBoneReg achieves robust, accurate, and modality-agnostic registration of bone surfaces, offering a promising solution for reliable cross-modal alignment in computer- and robot-assisted orthopedic surgery.

## 1. Introduction

Computer- and robot-assisted orthopedic surgery (CAOS) is a widely adopted discipline that has significantly enhanced accuracy and reproducibility in surgical interventions (Cleary and Peters, 2010), such as bone tumor resection (Evrard et al., 2022), total knee arthroplasty (TKA) (Sugano, 2003), and bone deformity correction (Baraza et al., 2020). A central prerequisite in many CAOS applications is the accurate registration between preoperative imaging data and surgery plans, including computed tomography (CT) and magnetic resonance imaging (MRI), with intraoperative data acquired from modalities such as optical tracking, ultrasound, X-ray and RGB-D imaging (Cleary and Peters, 2010; Scheufler et al., 2011; Kanlić et al., 2006; Picard et al., 2019). The accuracy of this registration process determines how precisely the surgery plan is realized intraoperatively, directly affecting navigation accuracy, instrument guidance, and surgical outcomes (Gautam et al., 2025; Wandvik et al., 2024; Lee et al., 2021). Achieving high accuracy across heterogeneous imaging modalities remains a key technical challenge and field of active research.

* Correspondence to: Lengghalde 5, Zurich, 8008, Zurich, Switzerland.
*E-mail address:* luohong.wu@balgrist.ch (L. Wu).

Existing registration approaches can be broadly categorized into image-based and surface-based methods, many designed for specific modality pairs (Kubicek et al., 2019; Zheng and Nolte, 2015). Image-based methods operate directly on volumetric or projection data without requiring explicit bone surface segmentation. Typical modality combinations include CT–CT (Tonetti et al., 2020; Park et al., 2017) and MRI–CT (Docquier et al., 2010; Pojskić et al., 2021) for 3D/3D registration, and CT–X-ray (Markelj et al., 2012; Otake et al., 2011; Penney et al., 1998) and MRI–X-ray (Markelj et al., 2012; Ku et al., 2023) for 3D/2D registration. More recently, registration of CT/MRI with ultrasound has gained considerable attention due to ultrasound's radiation-free and cost-effective nature (Hacihaliloglu et al., 2014; Roche et al., 2002; Penney et al., 2006; Pandey et al., 2018-05-26; Wein et al., 2007). Image-based methods generally optimize information-theoretic or intensity-based similarity measures such as mutual information in 3D/3D registration (Roche et al., 2002; Park et al., 2017; Pandey et al., 2018-05-26; Wein et al., 2007), or match digitally reconstructed radiographs (DRRs) to intraoperative X-rays in 3D/2D registration (Markelj et al., 2012). Despite promising results in specific use cases, image-based methods remain limited in robustness and generalization due to modality-specific artifacts, resolution disparities, and imaging device heterogeneity, constraining their clinical adoption (Markelj et al., 2012; Zheng and Nolte, 2015).

Surface-based registration mitigates these limitations by segmenting the anatomy of interest and representing its geometry as a point cloud, providing a compact and modality-agnostic representation. In open surgeries such as TKA, a common strategy involves manually sampling intraoperative point clouds of the target bone using a tracked pointer, which are then registered to point clouds derived from segmented CT or MRI data (Bae and Song, 2011; Anderson et al., 2005; Wilson et al., 2017). To automate sampling, RGB-D sensors have been employed as alternatives to the tracked pointer (Liebmann et al., 2024; Hu et al., 2021; Félix et al., 2021; Hu and Baena, 2022; Rodrigues et al., 2019). Direct surface sampling is effective in open surgeries with sufficient bone exposure but infeasible in minimally invasive procedures. Here, intraoperative imaging is utilized for surface reconstruction, such as intraoperative CT, X-rays (Grupp et al., 2020; Ku et al., 2025; Leung et al., 2010) and ultrasound (Wein et al., 2015; Salehi et al., 2017; Eppenhof et al., 2019; Guo et al., 2021; Ciganovic et al., 2018-04-05; Fanti et al., 2018; Li et al., 2023-05-12). Although surface-based registration facilitates cross-modality alignment, it remains challenging due to inherent differences in imaging physics that lead to noisy point clouds, varying spatial resolution and incomplete surface correspondences (Sotiras et al., 2013; Woo et al., 2014; Li et al., 2023-10-24; L. Wu, N.A. Cavalcanti, et al., 2025; Zhang et al., 2019; Liebmann et al., 2024; Hu et al., 2021). Registration is further complicated by geometric ambiguities often present in bone anatomy, leading to multiple equivalent solutions in the registration space (Ciganovic et al., 2018-04-05; Li et al., 2023-10-24). To mitigate these challenges, several strategies have been explored. Manual initialization followed by local refinement can achieve accurate results but prolongs procedures and requires expert intervention (Ma et al., 2017; Fanti et al., 2018; Li et al., 2023-05-12). General-purpose point cloud alignment methods such as RANSAC (Fischler and Bolles, 1981), fast global registration (FAST) (Zhou et al., 2016; Gsaxner et al., 2019), or principal component analysis (PCA)-based alignment can automate initialization but remain error-prone and sensitive to noise and partial data (Liu et al., 2022; Ao et al., 2025; Félix et al., 2021; Hu and Baena, 2022; Rodrigues et al., 2019; Liu et al., 2023; Hao et al., 2022; Zhang et al., 2019; Ao et al., 2025). Prior-driven methods, which exploit prior knowledge such as bone shape priors or estimated inter-modality transformations, improve robustness within specific settings but lack generalization across anatomies (Hacihaliloglu et al., 2014; Hu et al., 2021; Chan et al., 2021; Ciganovic et al., 2018-04-05). State-of-the-art (SOTA) learning-based methods have also shown promise for general surface registration by learning data-driven features from sufficiently large datasets. However, the lack of paired datasets limits the applicability of these powerful methods due to the difficulty in collecting real-world paired data (Unberath et al., 2021; Zhang et al., 2024). Consequently, robust and accurate automatic multimodal bone surface registration, especially for emerging intraoperative modalities such as ultrasound and RGB-D, remains an open research challenge. Instance-specific self-supervised methods are optimized for each new input instance without requiring a multi-instance dataset (Sun et al., 2020). Recent work has demonstrated their effectiveness in medical image analysis under limited-data settings (Chen et al., 2024). Based on instance-specific self-supervised learning, our proposed method is designed to address these challenges.

In this work, we introduce NeuralBoneReg, a modality-agnostic and instance-specific self-supervised framework for robust and accurate multimodal bone surface registration, using point clouds as an intermediate representation. NeuralBoneReg models bone geometry using a continuous implicit neural representation (INR) that enables accurate and differentiable point-to-surface distance evaluation across modalities. The framework comprises two main modules: a preoperative bone surface representation module (NeuralUDF) and an intraoperative registration module (NeuralReg). The NeuralUDF module employs a multi-layer perceptron (MLP) to learn an implicit unsigned distance field (UDF) from the preoperative point cloud. During the intraoperative stage, the NeuralReg module aligns the intraoperative point clouds within the learned UDF through parallel hypothesis optimization and a dedicated registration loss.

The main contributions of this study include:

- An instance-specific self-supervised approach for multimodal bone surface registration. We introduce the first instance-specific self-supervised, INR-based registration method, eliminating the need for large paired datasets or ground-truth transformations. This instance-wise training strategy enhances adaptability across anatomies and facilitates clinical translation.
- INR-based bone surface modeling. The proposed NeuralUDF module learns a continuous UDF from preoperative point clouds, enabling efficient and differentiable distance computation without nearest-neighbor search (Ciganovic et al., 2018-04-05; Castellani and Bartoli, 2020) or voxel interpolation (Salehi et al., 2017; Wein et al., 2015; Castellani and Bartoli, 2020). This design shifts computational load from the intraoperative to the preoperative stage, aligning with real-time requirements in CAOS.
- A parallel neural solver addressing surface ambiguities. To avoid convergence to incorrect local optima (Ciganovic et al., 2018-04-05), our NeuralReg module explores the SE(3) transformation space through multiple lightweight hypothesis generation heads sharing a common backbone. This shared-parameter, cooperative search improves performance over independent parallel optimization and accelerates global convergence.
- Comprehensive evaluation across three datasets. NeuralBoneReg and SOTA methods are evaluated on (i) an open-source CT–ultrasound dataset of the fibula and tibia (UltraBones100k) (L. Wu, N.A. Cavalcanti, et al., 2025), (ii) an open-source CT-RGB-D dataset of spinal vertebrae (SpineDepth) (Liebmann et al., 2021-08-27), and (iii) an in-house CT–ultrasound dataset of the femur and pelvis (UltraBonesHip). Ablation studies are conducted to quantify the contribution of each proposed component.

- Release of the UltraBonesHip dataset. To support future research, we publicly release the UltraBonesHip dataset, comprising paired CT and ultrasound volumes of human ex-vivo specimens with annotated bone segmentations and CT–ultrasound ground-truth transformations.

To ensure reproducibility of the results, the code and all data have been made publicly available.[1]

## 2. Related work

In this section, we present prior work on surface-based multimodal bone surface registration as well as the SOTA in deep learning-based point cloud registration.

**Surface-Based Multimodal Bone Surface Registration.** Surface-based methods represent anatomical structures as point clouds extracted from preoperative and intraoperative imaging data. A typical registration workflow consists of coarse global initialization followed by local refinement, most commonly via iterative closest point (ICP) (Besl and McKay, 1992). Early studies achieved global initialization through manually selected correspondences. For instance, Ma et al. (2017), Fanti et al. (2018), and Li et al. (2023-05-12) registered point clouds derived from CT and ultrasound data of the spine and lower extremities using manual landmark selection and subsequent ICP refinement, achieving mean target registration errors of $< 2$ mm. While accurate, such approaches are labor-intensive, require expert knowledge, and erroneous point sampling requires re-registration, prolonging the procedure by up to 20 minutes (Nottmeier and Crosby, 2007; Bae and Song, 2011). To reduce manual effort, fully automatic global initialization approaches have been proposed that typically rely on feature-based or probabilistic methods such as RANSAC (Fischler and Bolles, 1981), FAST (Zhou et al., 2016; Gsaxner et al., 2019), or PCA-based alignment. Ao et al. applied fast point feature histograms (FPFH)-based RANSAC and FAST followed by ICP for CT–ultrasound spine registration in a robotic surgery simulator (Ao et al., 2025). Similar RANSAC–ICP pipelines have been employed for CT/MRI to RGB-D registration in TKA (Félix et al., 2021; Rodrigues et al., 2019), and for femoral registration between optical scanner and RGB-D scans (Hu and Baena, 2022). FPFH-based RANSAC with ICP refinement has also been used for CT to intraoral scanner registration of alveolar bone (Liu et al., 2023) and crown–root–bone alignment (Hao et al., 2022). Liu et al. performed CT to ultrasound registration of lumbar vertebrae by PCA-based alignment of centroids and orientation, followed by ICP refinement (Liu et al., 2022). Although these approaches achieve reliable alignment under controlled conditions and moderate overlap (error $\approx 2°/1$ mm (Ao et al., 2025)), they remain sensitive to noise, outliers, and incomplete surfaces in real-world datasets (error $\approx 5°/6$ mm (Félix et al., 2021; Rodrigues et al., 2019)). To improve robustness, prior knowledge and geometric constraints have been incorporated into automated pipelines. Hacihaliloglu et al. aligned ultrasound-derived lumbar spine reconstructions with a preoperative statistical shape-and-pose model, incorporating ultrasound probe position as a prior for the initialization (Hacihaliloglu et al., 2014). Hu et al. registered intraoperative RGB-D-derived point clouds of the femur to preoperative MRI-derived point clouds using a prior-correspondence-based global stage followed by ICP refinement (Hu et al., 2021). Ciganovic et al. (2018-04-05) and Chan et al. (2021) exploited PCA-based initialization that embeds geometric shape and pose priors for MRI–ultrasound and CT–ultrasound registration, followed by ICP refinement. A similar prior-driven strategy has been employed for CT–ultrasound registration of the pelvis based on known geometry symmetry and main bone axis (Foroughi et al., 2008). Although prior-driven approaches achieve strong performance on real-world datasets (error $\approx 1°/\approx 1$ mm (Ciganovic et al., 2018-04-05; Chan et al., 2021)), their strong dependence on anatomy-specific priors and initialization hinders generalization across different bone anatomies.

**Deep Learning-based Point Cloud Registration.** The growing availability of large-scale, annotated datasets has enabled deep learning-based approaches for point cloud registration. These methods have emerged as powerful alternatives to traditional approaches, offering greater robustness, scalability, and the ability to learn data-driven priors (Zhang et al., 2024, 2020). They can be categorized into three main categories: neural feature extractors for correspondence matching, initialization models for global alignment, and end-to-end registration models for regressing the final transformation parameters (Zhang et al., 2024). Unlike handcrafted descriptors such as FPFH, neural feature extractors provide robust correspondence estimation under noise, clutter, or partial overlap by learning discriminative, rotation-invariant representations directly from raw point sets, including FCGF (Choy et al., 2019), 3DSmoothNet (Gojcic et al., 2019), SpineNet (Ao et al., 2021), 3DMatch (Zeng et al., 2017), YOHO (Wang et al., 2022). Directly relevant to our partial-to-complete registration task, Huang et al. proposed Predator to improve correspondence extraction by predicting overlapping regions and keypoint saliency under low-overlap conditions (Huang et al., 2021). Based on the deep features, a traditional coarse-to-fine registration workflow can then be applied to estimate the final transformation parameters. Neural feature extractors have shown to outperform handcrafted feature descriptors such as FPFH, achieving mean rotation/translation error of 1.74°/19 mm on ModelNet and 2.03°/60 mm on 3DMatch (Huang et al., 2021). Although effective, the optimal transformation estimator requires dataset-specific parameter tuning. To overcome this limitation, deep networks such as 3DRegNet (Pais et al., 2020) and DeepGMR (Yuan et al., 2020) have been proposed to predict global initialization followed by a conventional local refinement method. For instance, Liebmann et al. proposed a supervised network where the global alignment was learned using an MLP that learns a PCA-based initialization for aligning RGB-D-derived lumbar spine point clouds (L1–L5 posterior surface) with preoperative CT bone models (Liebmann et al., 2024), achieving an average mean target registration error of 2.72 mm. Recent studies further streamlined registration by jointly solving global and local alignment in an end-to-end network, such as DCP (Wang and Solomon, 2019), DeepVCP (Lu et al., 2019), and FINet (Xu et al., 2022). Qin et al. introduced GeoTransformer (Qin et al., 2023), which encodes pairwise distances and triplet-wise angles between superpoints to learn transformation-invariant features for robust correspondence matching. The correspondences are then processed by a differentiable singular value decomposition (SVD) layer for transformation regression, achieving rotation/translation errors of 0.27°/68 mm on KITTI and 1.74°/19 mm on ModelNet (Qin et al., 2023). Although end-to-end methods achieve strong performance, they rely on ground-truth transformations for supervision. Self-supervised methods, such as PPF-FoldNet (Deng et al., 2018) and DiReg (Löwens et al., 2024), optimize geometric consistency objectives (e.g., Earth Mover's Distance or cycle consistency) without requiring explicit labels, achieving comparable accuracy as supervised networks (3.578°/67 mm on ModelNet and 2.951°/86 mm on 3DMatch). However, in contrast to our approach, these methods still require pretraining on large datasets of paired point clouds, which are rarely available for multimodal bone surfaces (Unberath et al., 2021).

We hypothesize that limited dataset size in our task is the primary bottleneck for the performance of deep learning-based methods. To confirm this, our experiments will benchmark Predator (Huang et al., 2021) and GeoTransformer (Qin et al., 2023) on all datasets.

## 3. Methods

NeuralBoneReg is a surface-based approach designed to register preoperative imaging data (e.g. CT or MRI) with intraoperative imaging data (e.g. ultrasound or RGB-D). We begin by presenting the problem formulation in Section 3.1. Our proposed method is detailed in Section 3.2, comprising the NeuralUDF module in Section 3.2.1 and the NeuralReg module in Section 3.2.2.

[1] https://neuralbonereg.github.io/

### 3.1. Problem formulation

We assume that a segmented bone surface from preoperative imaging modality is available, and denote it as $\mathcal{S}$. A point cloud $\mathcal{C}^{\text{raw}}$ of size $M^{\text{raw}}$ is uniformly and randomly sampled from $\mathcal{S}$, represented as

$$\mathcal{C}^{\text{raw}} = \left\{ \mathbf{c}_i^{\text{raw}} = \left[x_i, y_i, z_i\right]^T \in \mathbb{R}^3 \mid i \in [1, \ldots, M^{\text{raw}}] \right\} \tag{1}$$

The spatial scale of $\mathcal{C}^{\text{raw}}$ is defined as the diagonal length of the 3-dimensional bounding box:

$$\ell_{\mathcal{C}}^{\text{raw}} = \sqrt{\left(\max_i(x_i) - \min_i(x_i)\right)^2 + \left(\max_i(y_i) - \min_i(y_i)\right)^2 + \left(\max_i(z_i) - \min_i(z_i)\right)^2} \tag{2}$$

The centroid of $\mathcal{C}^{\text{raw}}$ is computed as:

$$\overline{\mathcal{C}^{\text{raw}}} = \begin{bmatrix} \bar{x} \\ \bar{y} \\ \bar{z} \end{bmatrix} = \frac{1}{M^{\text{raw}}} \sum_{i=1}^{M^{\text{raw}}} \mathbf{c}_i^{\text{raw}} \tag{3}$$

Following prior work (L. Wu, M. Seibold, et al., 2026; Chen et al., 2024; Ma et al., 2020), we apply spatial normalization to map $\mathcal{C}^{\text{raw}}$ into the normalized cube $(-1,1)^3$ using the transformation:

$$\frac{\mathbf{c}_i^{\text{raw}} - \overline{\mathcal{C}^{\text{raw}}}}{\ell_{\mathcal{C}}^{\text{raw}}} \tag{4}$$

We then discretize the space into a voxel grid and perform downsampling to achieve uniform point density. The resulting preoperative point cloud is denoted as:

$$\mathcal{C} = \left\{ \mathbf{c}_i \in (-1,1)^3 \mid i \in [1, \ldots, M],\ M \le M^{\text{raw}} \right\} \tag{5}$$

The intraoperative imaging data provide a partial and noisy observation of the target bone surface $\mathcal{S}'$, acquired through modalities such as point-sampling, ultrasound, or RGB-D imaging. For point-sampling with a tracked pointer, the intraoperative point cloud $\mathcal{U}^{\text{raw}}$ is directly constructed from the sensor measurements (Bae and Song, 2011). For ultrasound and RGB-D imaging, $\mathcal{U}^{\text{raw}}$ is generated via 2D segmentation followed by 3D reconstruction using recorded pose information (L. Wu, N.A. Cavalcanti, et al., 2025; Liebmann et al., 2024). The resulting intraoperative point cloud $\mathcal{U}^{\text{raw}}$ is denoted as:

$$\mathcal{U}^{\text{raw}} = \left\{ \mathbf{u}_i^{\text{raw}} = \left[x_i, y_i, z_i\right]^T \in \mathbb{R}^3 \mid i \in [1, \ldots, N^{\text{raw}}] \right\} \tag{6}$$

Analogous to $\mathcal{C}^{\text{raw}}$, the spatial scale $\ell_{\mathcal{U}}^{\text{raw}}$ and the shape centroid $\overline{\mathcal{U}^{\text{raw}}}$ are computed. As intraoperative data typically covers only a partial surface, $\ell_{\mathcal{U}}^{\text{raw}} < \ell_{\mathcal{C}}^{\text{raw}}$ generally holds. $\mathcal{U}^{\text{raw}}$ is normalized using $\ell_{\mathcal{C}}^{\text{raw}}$ to ensure scale consistency with $\mathcal{C}$, followed by voxel discretization and downsampling. The resulting intraoperative point cloud $\mathcal{U}$ is denoted as:

$$\mathcal{U} = \left\{ \mathbf{u}_i \in (-1,1)^3 \mid i \in [1, \ldots, N],\ N \le N^{\text{raw}} \right\} \tag{7}$$

In summary, the registration task is defined as estimating the 6-Degrees-of-Freedom (6-DoF) rigid transformation $\mathbf{T}_{\text{C}\leftarrow\text{U}} = [\mathbf{R} \mid \mathbf{t}] \in \text{SE}(3)$ that aligns the intraoperative point cloud $\mathcal{U}$ with the preoperative point cloud $\mathcal{C}$ by minimizing the following objective function:

$$\min_{\mathbf{R}\in\text{SO}(3),\ \mathbf{t}\in\mathbb{R}^3} \frac{1}{N} \sum_{i=1}^{N} \left\| \mathbf{c}_{\pi(i)} - (\mathbf{R}\mathbf{u}_i + \mathbf{t}) \right\|_2 \quad \text{s.t. } \mathbf{c}_{\pi(i)} \in \mathcal{C},\ \mathbf{u}_i \in \mathcal{C}. \tag{8}$$

where $\pi(i) = \arg\min_{j=1,\ldots,M} \left\| \mathbf{c}_j - (\mathbf{R}\cdot\mathbf{u}_i + \mathbf{t}) \right\|$ denotes the index of the nearest neighbor of the transformed point $(\mathbf{R}\mathbf{u}_i + \mathbf{t})$ in $\mathcal{C}$.

### 3.2. Proposed method

Aligned with the CAOS workflow, comprising preoperative and intraoperative stages, NeuralBoneReg includes two corresponding modules: preoperative NeuralUDF and intraoperative NeuralReg, as illustrated in Fig. 1. In the preoperative stage, a neural UDF is learned from the preoperative point cloud $\mathcal{C}$. During the intraoperative stage, NeuralReg generates multiple registration transformation hypotheses in parallel to align the intraoperative point cloud $\mathcal{U}$ with the learned neural UDF, solving both global initialization and local refinement efficiently. Each hypothesis is evaluated based on the mean distance within the learned UDF, and the hypothesis yielding the lowest mean distance is selected as the final result.

#### 3.2.1. NeuralUDF

Eq. (8) corresponds to the objective optimized in conventional point-cloud-based registration, such as in a RANSAC–ICP pipeline. Minimizing this objective requires nearest-neighbor search, which is non-differentiable and computationally expensive, particularly in iterative optimization pipelines. To overcome this challenge, Wein et al. (2015) fused ultrasound data with a fixed-resolution signed distance field (SDF) derived from preoperative CT. The fixed-resolution SDFs represent the soft tissue layer between bone and skin, providing gradients that enable simulated probe-induced compression during intensity-based local refinement. Building on this idea, we propose to learn a neural UDF from the preoperative point cloud $\mathcal{C}$, which offers several advantages. The learned UDF supports direct distance queries from arbitrary 3D points, eliminating the need for explicit, non-differentiable nearest-neighbor search. In addition, unlike fixed-resolution distance volumes, a neural UDF provides effectively arbitrary resolution and smooth, differentiable objectives that enable efficient gradient-based optimization. As we will show in Section 5, fixed-resolution representations degrade when the target anatomy spans larger scales (e.g., pelvis vs. vertebra).

Several methods have been proposed for learning a neural implicit representation from a point cloud (L. Wu, M. Seibold, et al., 2026; Chen et al., 2023; Ma et al., 2020), all of which can be applied to our NeuralUDF module. Among these methods, UltraBoneUDF has shown effectiveness in learning neural UDF for both partial and complete bone anatomies (L. Wu, M. Seibold, et al., 2026). The network learns to approximate the unsigned distance of any spatial location to the underlying surface by minimizing a tangent-plane consistency loss and a coarse global-shape regularizer. This yields a smooth, differentiable distance field that implicitly encodes the bone surface and supports direct geometric reasoning. In our work, we apply UltraBoneUDF with default parameters. For details of the training procedure, including the architecture and loss function $\mathcal{L}_{\text{UDF}}$, we refer to L. Wu, M. Seibold, et al. (2026).

Though more time-consuming to train, the NeuralUDF module is computed preoperatively, offloading nearest-neighbor computations and allowing faster intraoperative optimization. After training, the UDF defines a mapping $d(\mathbf{q}_i) : \mathbb{R}^3 \to \mathbb{R}_{\ge 0}$, assigning each query point $\mathbf{q}_i \in \mathbb{R}^3$ a non-negative value representing its predicted distance to $\mathcal{S}$. The frozen NeuralUDF is used for distance queries in the NeuralReg module.

#### 3.2.2. NeuralReg

Leveraging the learned UDF, the registration task is reformulated as minimizing the mean UDF value over the transformed intraoperative point cloud:

$$\min_{[\mathbf{R}|t]\in\text{SE}(3)} \frac{1}{N} \sum_{i=1}^{N} d(\mathbf{R}\mathbf{u}_i + \mathbf{t}) \tag{9}$$

In contrast to Eq. (8), the proposed formulation is continuous and differentiable with respect to $[\mathbf{R} \mid \mathbf{t}]$, enabling direct gradient-based optimization.

However, geometric symmetries in $\mathcal{S}$ induce numerous local minima in the objective, making direct optimization of the 6-DoF transformation, whether deterministic or stochastic, unreliable. As shown later in our ablation studies, this naive approach often converges to suboptimal solutions, highlighting the need for a more robust optimization strategy.

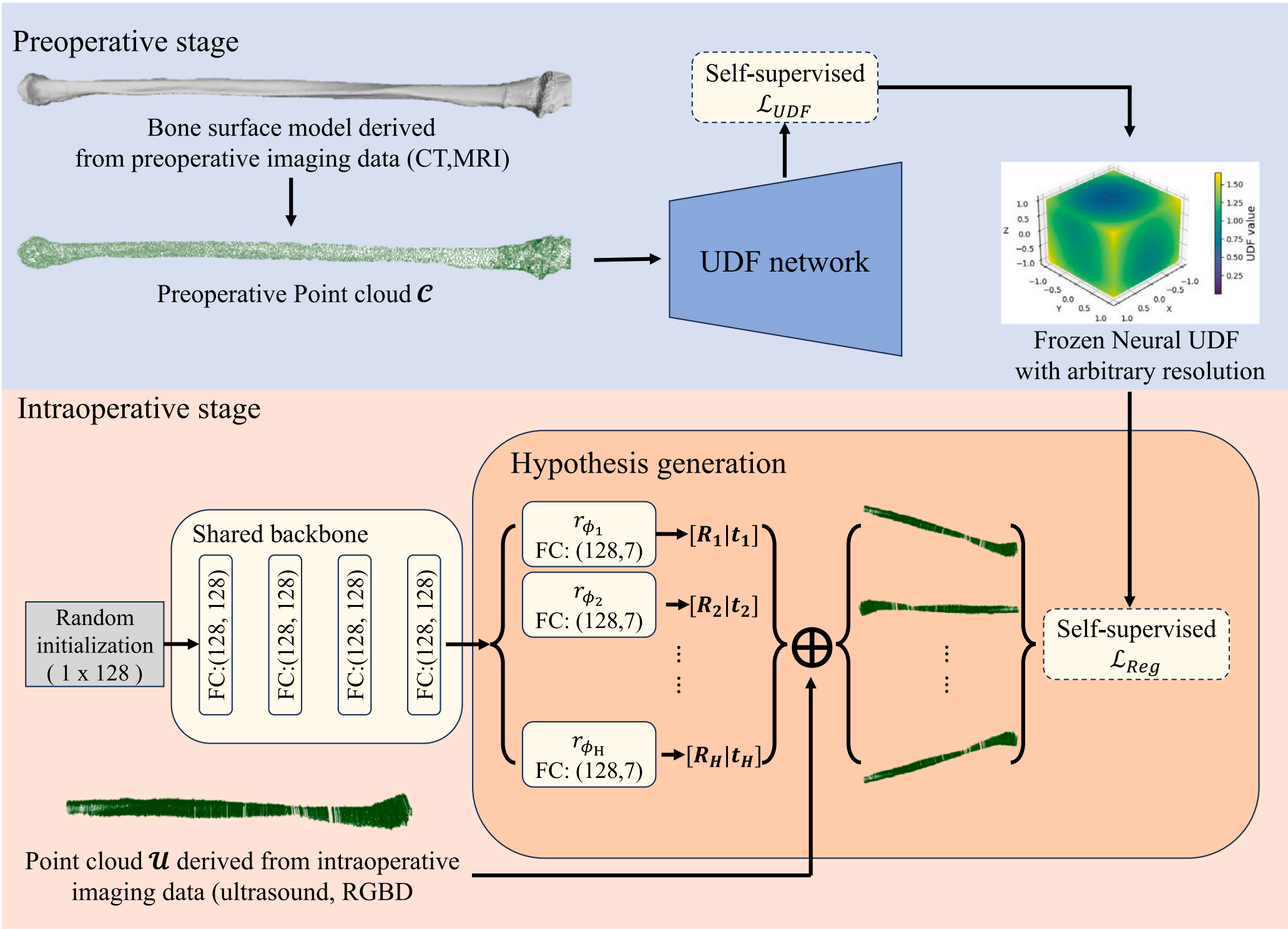

**Fig. 1.** Network architecture of the two-stage approach of NeuralBoneReg. In the preoperative stage, the NeuralUDF module learns an implicit neural representation of the target bone surface from the preoperative data $\mathcal{C}$. During the intraoperative stage, the NeuralReg module leverages multiple lightweight neural heads built on a randomly initialized shared backbone to efficiently generate registration hypotheses aligning the intraoperative data $\mathcal{U}$ with the preoperative data $\mathcal{C}$. FC($\mathrm{dim_{in}}$, $\mathrm{dim_{out}}$) denotes a fully connected layer with input dimension $\mathrm{dim_{in}}$ and output dimension $\mathrm{dim_{out}}$, followed by a Tanh activation.

A strategy to mitigate such ambiguities is to explore multiple candidate solutions in parallel. For instance, Bishop's Mixture Density Networks predict the parameters of a mixture probability distribution, allowing the modeling of multiple plausible outputs for a shared input (Bishop, 1994). Inspired by this concept, our registration network is designed to generate and evaluate multiple transformation hypotheses in parallel. Given the input point cloud $\mathcal{U}$ and the pretrained UDF, the key idea is to leverage multiple hypothesis generation heads to explore different regions of SE(3) simultaneously. This parallel hypothesis generation increases the likelihood of finding the correct alignment without relying on sequential local refinements, while the shared backbone allows information exchange between hypotheses to improve overall convergence stability. The depth of the shared backbone determines how effectively information is shared across heads, while the number of heads controls the diversity of explored transformations. Section 4.5 presents ablation studies analyzing the influence of both factors on registration performance.

At the architectural level, NeuralReg comprises a shared MLP backbone $f_\theta$ and $H$ hypothesis generation heads $\{r_{\phi_h}\}_{h=1}^{H}$, as illustrated in Fig. 1. Given a randomly sampled initialization vector $\bar{\mathbf{x}} \in (-1,1)^{128}$, the shared backbone maps a latent feature $f_\theta(\bar{\mathbf{x}}) \in (-1,1)^{128}$, that serves as a common conditional signal for all heads. During training, $\bar{\mathbf{x}}$ is held fixed over the overall training for a single input point cloud while the backbone parameters $\theta$ are optimized jointly. Conditioned on $d$, each head implements a mapping $r_\phi(\mathcal{U}, f_\theta(\bar{\mathbf{x}}) \mid d) \rightarrow \mathbb{R}^3 \times \mathbb{R}^4$ that outputs a translation vector $\mathbf{t} \in \mathbb{R}^3$ and a quaternion $\eta \in \mathbb{R}^4$ ( $\|\boldsymbol{\eta}\|_2 = 1$). The quaternion $\eta$ is converted into a rotation matrix $\mathbf{R} \in$ SO(3), which is then combined with $\mathbf{t}$ to define a rigid transformation hypothesis $[\mathbf{R} \mid \mathbf{t}] \in \mathrm{SE}(3)$. The alignment quality of each transformation $[\mathbf{R} \mid \mathbf{t}]$ is evaluated through Eq. (9), which serves as the training objective for the network. To efficiently exploit GPU parallelism, each hypothesis generation head is implemented as a lightweight fully connected layer followed by a tanh activation ($\tanh(x) = \frac{e^x - e^{-x}}{e^x + e^{-x}}$), chosen for its normalized output range $\tanh(x) \in (-1,1)$. To ensure valid unit-norm representation, each output quaternion is subsequently normalized such that $\|\boldsymbol{\eta}\|_2 = 1$. While all heads share the backbone parameters $\theta$, they maintain distinct parameter sets $\{\phi_h\}_{h=1}^{H}$.

The registration network is trained in an instance-specific self-supervised manner using the following loss function $\mathcal{L}_{\mathrm{Reg}}$:

$$\mathcal{L}_{\mathrm{Reg}} = \frac{1}{H} \sum_{h \in [1,H]} \left[ \frac{1}{B} \sum_{i \in [1,B]} d(\mathbf{R}_h \mathbf{u}_i + \mathbf{t}_h) \right]. \tag{10}$$

where $B$ denotes the batch size. For hypotheses close to the solution, a large $B$ provides higher spatial resolution and improves precision. However, for clearly incorrect hypotheses, such high resolution is unnecessary and increases computation and runtime. To improve temporal and spatial efficiency, we adopt a coarse-to-fine strategy. In the coarse stage, all hypotheses are evaluated using a sparse point cloud with $B_{\mathrm{sparse}}$ points. The top-$K$ hypotheses are then re-evaluated in the fine stage using a denser point cloud with $B$ points. Ablation studies are conducted to evaluate the influence of different choices of these values in Section 4. In addition, in CAOS, aligning centroids of $\mathcal{U}$ and $\mathcal{C}$ provides a rough estimate of the registration translation, while the main challenge lies in estimating the rotation. Therefore, the translation outputs of all heads are randomly initialized, whereas the rotation outputs are initialized by farthest point sampling (FPS) in the SO(3) space.

During inference, the hypothesis yielding the lowest mean UDF value is selected as the final solution.

## 4. Experiments

To evaluate NeuralBoneReg, we perform a comprehensive comparison against existing SOTA registration methods. Section 4.1 introduces the evaluation datasets, followed by implementation details in Section 4.2. The SOTA approaches are described in Section 4.3, and the evaluation metrics are summarized in Section 4.4. Section 4.5 describes the ablation studies.

**Table 1**
Demographic characteristics of the subjects in UltraBonesHip. Each subject includes pelvis and proximal femur.

| Specimen No. | Height (cm) | Weight (kg) | BMI | Age | Gender | Arthritis |
|---|---|---|---|---|---|---|
| 1 | 188 | 98 | 27.86 | 66 | Male | History of falls |
| 2 | 180 | 93 | 28.59 | 66 | Male | History of falls |
| 3 | 168 | 75 | 26.63 | 72 | Female | History of falls |
| 4 | 165 | 76 | 27.95 | 71 | Female | None specified |
| 5 | 178 | 77 | 24.39 | 49 | Male | None specified |

### 4.1. Datasets

Given the growing body of research in CT–ultrasound registration, we conduct experiments on two CT–ultrasound datasets. To further assess the generalizability of NeuralBoneReg beyond ultrasound, an additional CT-RGB-D dataset is included. Together, these datasets cover a range of anatomies such as the fibula, tibia, femur, pelvis, and lumbar spine.

#### 4.1.1. UltraBones100k

UltraBones100k is an open-source dataset collected by clinical experts under realistic clinical setup conditions. It comprises CT-derived cortical bone surfaces and tracked 3D ultrasound recordings from 14 human ex-vivo lower-extremity specimens, including the tibia and fibula (L. Wu, N.A. Cavalcanti, et al., 2025). For each subject, multiple ultrasound scans are available per bone. A pretrained segmentation model is provided to generate bone segmentation masks from the ultrasound data. Following the preprocessing steps described in Section 3.1, we generate $\mathcal{C}$ and $\mathcal{U}$ for each bone of each subject, resulting in 28 paired point clouds (14 subjects × 2 bones). As the bones were tracked and the ultrasound probe was calibrated in the same coordinate system during dataset acquisition, $\mathcal{C}$ and $\mathcal{U}$ are already aligned (L. Wu, N.A. Cavalcanti, et al., 2025). All preprocessed point clouds ($\mathcal{C}$, $\mathcal{U}$) are available for downloading via our repository.[2]

#### 4.1.2. UltraBonesHip

To expand the evaluation to cover more anatomies, we collected an additional CT–ultrasound dataset of the hip region, including the femur and pelvis. Data acquisition followed the UltraBones100k protocol with CT-derived cortical bone surface segmentation (L. Wu, N.A. Cavalcanti, et al., 2025) and involved 5 human ex-vivo specimens (see Table 1). Each specimen was scanned along the coronal and axial directions. Ethical approval for conducting this study was obtained from the Zurich Cantonal Ethical Committee (BASEC Nr. 2023-01652). We directly applied the pretrained segmentation model from UltraBones100k without any fine-tuning (L. Wu, N.A. Cavalcanti, et al., 2025). Unlike the fibula and tibia, which are close to the skin, hip anatomies are deeper and more challenging for ultrasound imaging, especially in specimens with higher BMI and thicker soft tissue over the bone, resulting in larger cross-specimen variation than the UltraBones100k dataset. The limited number of specimens ($n = 5$) further makes this dataset more challenging for SOTA learning-based methods. The processing steps described in Section 3.1 were applied to generate $\mathcal{C}$ and $\mathcal{U}$, resulting in 5 subjects × 3 anatomical structures (left femur, right femur, and pelvis) = 15 pairs of point clouds. The raw data, ground-truth transformations, and preprocessed point clouds are publicly available via our repository.

[2] https://neuralbonereg.github.io/

#### 4.1.3. SpineDepth

The SpineDepth dataset comprises CT and RGB-D data of lumbar spines from 10 human ex-vivo subjects (Liebmann et al., 2021-08-27). For each subject, CT-derived cortical bone surfaces are available for each lumbar vertebra (L1 to L5). In addition, RGB-D recordings of the exposed spine were acquired during the ex-vivo surgery using a ZED Mini camera (Stereolabs Inc., San Francisco, CA, USA) from two viewpoints, and the ground-truth inter-modality transformations were obtained through optical tracking and camera calibration (Liebmann et al., 2021-08-27). Following prior work (Liebmann et al., 2024), we exclude subjects #1 and #10 due to limited anatomical exposure. For each of the remaining 8 subjects, RGB-D scans of the same vertebral level and viewpoint were combined, yielding $8 \times 5 \times 2 = 80$ $\mathcal{C}$–$\mathcal{U}$ pairs. We make preprocessed point clouds publicly available via our repository.

### 4.2. Implementation details

In our experiments, we set $|\mathcal{C}| = M = 40{,}000$ and $|\mathcal{U}| = N = 40{,}000$. For the NeuralUDF module, we apply UltraBoneUDF with default parameters (L. Wu, M. Seibold, et al., 2026). For NeuralReg, based on the empirical results in Section 5.2, the number of hypothesis generation heads is set to $H = 5{,}000$, with batch size $B = 3{,}000$, $B_{\text{sparse}} = 32$, and $K = 64$. The training was performed using the Adam optimizer (learning rate: 0.001, momentum: 0.9) for 1000 iterations.

The entire framework was implemented in PyTorch v2.7.0 with CUDA 12.8 support. Differential operations, including quaternion-rotation matrix conversions, were implemented using PyTorch3D v0.7.8 (Ravi et al., 2020). The code is publicly available via our repository.

### 4.3. Baseline methods

In clinical practice, preoperative and intraoperative data are not aligned, resulting in unknown initial pose discrepancies. To simulate this scenario, we sampled a random transformation for $\mathcal{U}$, with rotations in $[-180^\circ, 180^\circ]$ and translations in $[-500, 500]$ mm along each axis. Each registration method was tasked with recovering the inverse transformation $\mathbf{T}_{\text{per}}^{-1}$. To ensure robustness, each experiment was repeated 20 times with independent random initializations per $\mathcal{C}$-$\mathcal{U}$ pair.

In the experiments, we compare the performance of NeuralBoneReg against the following baseline methods:

- **Traditional Methods.** We evaluate three commonly used two-stage approaches combining coarse global initialization with local refinement: PCA+ICP (Ciganovic et al., 2018-04-05; Liu et al., 2022; Foroughi et al., 2008), RANSAC+ICP (Félix et al., 2021; Hu and Baena, 2022; Rodrigues et al., 2019; Liu et al., 2023; Ao et al., 2025; Hao et al., 2022). Following prior work, FPFH features are used for RANSAC. In PCA+ICP, we improve global initialization robustness by sampling multiple rotations around the first three principal axes (Ciganovic et al., 2018-04-05). For ICP, both point-to-point and point-to-plane variants are evaluated, and the better result is reported for each dataset. This policy is applied to all ICP-based experiments. Implementations are based on Open3D.[3]
- **Learning-based Methods.** To benchmark against learning-based approaches for point cloud registration (Zhang et al., 2024), we include four SOTA networks: Predator (Huang et al., 2021),[4] GeoTransformer (Qin et al., 2023),[5] CASTv2 (Huang et al., 2026)[6] and DiReg (Löwens et al., 2024).[7] Their official implementation with default parameters is applied to each evaluation dataset using specimen-wise leave-one-out cross-validation. Reported results represent the mean performance across all validation splits.
- **Pseudo Ground-Truth.** To estimate the upper performance bound, we apply ICP initialized with the ground-truth transformation for each $\mathcal{C}$–$\mathcal{U}$ pair, and define the results as the pseudo ground-truth.

[3] https://www.open3d.org/docs/release/tutorial/pipelines/global_registration.html
[4] https://github.com/prs-eth/OverlapPredator
[5] https://github.com/qinzheng93/GeoTransformer
[6] https://github.com/RenlangHuang/CASTv2

### 4.4. Evaluation metrics

For evaluation on the transformation level, we report the relative rotation error (RRE) and relative translation error (RTE) (Zhang et al., 2024; Huang et al., 2021). Given the ground-truth transformation $\mathbf{T}_{\text{GT}} = [\mathbf{R}_{\text{GT}} \mid \mathbf{t}_{\text{GT}}]$ and the estimated transformation $\mathbf{T}_{\text{est}} = [\mathbf{R}_{\text{est}} \mid \mathbf{t}_{\text{est}}]$, they are defined as

$$\text{RRE}(\mathbf{R}_{\text{est}}) = \arccos\left(\frac{\text{Tr}(\mathbf{R}_{\text{GT}}^{\top} \cdot \mathbf{R}_{\text{est}}) - 1}{2}\right) \tag{11}$$

and

$$\text{RTE}(\mathbf{t}_{\text{est}}) = \left\|\mathbf{t}_{\text{est}} - \mathbf{t}_{\text{GT}}\right\|_2 . \tag{12}$$

Following prior work (Weber et al., 2024), we also report registration recall as the percentage of experiments with RTE $< 4$ mm and RRE $< 5\,°$.

To evaluate registration quality at the point cloud level, we report the Chamfer Distance (CD) and the 95% Hausdorff Distance (HD95) between the registered intraoperative point cloud and the preoperative point cloud. We further report recall curves over different CD thresholds to characterize the performance distribution.

### 4.5. Ablation studies

To assess the contribution of individual design choices in NeuralBoneReg, we conducted ablation studies focusing on the following aspects.

**Coarse-to-fine Configuration.** To evaluate how different parameter values affect the coarse-to-fine strategy, we set $H \in [1000, 5000, 10000]$, $B_{\text{sparse}} \in [32, 64, 128]$, and $K \in [0, 32, 64, 128]$ on the UltraBones100k dataset. For each configuration, we also report the number of parameters, FLOPs per iteration, and runtime, in addition to the evaluation metrics.

**The Optimization Approach (Neural vs Traditional).** NeuralBoneReg integrates a neural distance field with a neural optimization module. To systematically assess the contribution of these components, we compare neural, hybrid, and traditional variants. Moreover, we analyze combinations of implicit and grid-based distance field representations with either neural or traditional optimizers across all datasets:

- NeuralUDF+BFGS: To quantify the performance gain over a traditional gradient-based optimizer, we employ Broyden–Fletcher–Goldfarb–Shanno (BFGS) to optimize Eq. (9). BFGS, a quasi-Newton method that approximates the Hessian for efficient optimization of smooth functions, has been used in CT–ultrasound bone surface registration (Fletcher, 2000; Winter et al., 2008). The implementation provided in the SciPy library is used.[8]
- NeuralUDF+DE: The BFGS variant optimizes Eq. (9) by exploiting gradient information from the initialization. However, the pronounced symmetry of most human bones results in an optimization landscape with numerous local minima, increasing the likelihood of incorrect convergence. As an alternative to gradient-based methods, the differential evolution (DE) algorithm has been widely employed for such highly non-convex optimization problems (Winter et al., 2008; Storn and Price, 1997; Pant et al., 2020). The DE algorithm mitigates local minima by stochastically exploring the search space in parallel through a population of candidate solutions, and converges when the population stabilizes. The implementation provided in the SciPy library is used.[9]
- GridVolume+NeuralReg: To quantify the performance gain achieved by an implicit neural representation, a discrete fixed-resolution distance volume is constructed from the preoperative point cloud $\mathcal{C}$, where each voxel stores the distance to its nearest neighbor in $\mathcal{C}$ (Salehi et al., 2017; Wein et al., 2015; Castellani and Bartoli, 2020). Distances at arbitrary query points are obtained by linear interpolation. To balance resolution and runtime, we set the grid resolution to $512^3$ across all datasets and use NeuralReg for optimization.
- GridVolume+BFGS: This variant combines conventional grid-based point queries and a traditional gradient-based optimizer.

**Grid Resolutions.** Grid-based methods employ fixed-resolution distance volumes. While this representation may adequately capture smaller anatomies, larger bones require substantially higher grid resolutions to achieve comparable surface detail. To investigate the effect of grid resolution, we evaluate the GridVolume variant using volumes of $[128^3, 256^3, 512^3]$ on the UltraBonesHip dataset (largest anatomy) and the SpineDepth dataset (smallest anatomy).

**Shared Backbone for Joint Learning.** Sharing backbone parameters allows gradient updates to propagate across heads, potentially accelerating convergence over independent optimization. The number of shared parameters is therefore expected to influence the effectiveness of this propagation mechanism. To quantify this effect, we vary the backbone depth within the range $[0, 1, 2, 3, 4, 5, 6]$ (0 indicating no parameter sharing) and evaluate the variants on the SpineDepth dataset, which contains the largest number of $\mathcal{C}$–$\mathcal{U}$ pairs among the three datasets.

**Robustness to Noise.** Intraoperative point clouds $\mathcal{U}$ can be corrupted by noise from multiple sources, including 2D segmentation errors and 3D tracking noise. To assess the robustness of our method, we add Gaussian noise $\mathcal{N}(0, \sigma^2)$ to the ultrasound-derived point clouds in UltraBones100k, where $\sigma \in [0.1, 0.5, 1, 2, 3, 4, 5]$.

## 5. Results

### 5.1. Main results

As shown in Table 2, compared with SOTA methods, our method achieves competitive performance on the UltraBones100k dataset, with an RRE of $1.83 \pm 1.30°$, an RTE of $2.02 \pm 1.30$ mm, an RR of 0.89, a CD of $0.82 \pm 0.12$ mm, and an HD95 of $2.06 \pm 0.36$ mm. Similarly, on the SpineDepth dataset (Table 4), our method achieves competitive results, with an RRE of $3.78 \pm 19.34°$, an RTE of $2.80 \pm 3.75$ mm, an RR of 0.84, a CD of $1.78 \pm 1.61$ mm, and an HD95 of $4.26 \pm 4.58$ mm. On the UltraBonesHip dataset (Table 3), the performance of SOTA methods drops significantly, whereas our method maintains similar performance, with an RRE of $1.90 \pm 1.56°$, an RTE of $2.21 \pm 0.86$ mm, an RR of 0.88, a CD of $2.50 \pm 1.08$ mm, and an HD95 of $8.97 \pm 4.08$ mm. In all cases, the registration accuracy remained within approximately $1°/1$ mm of the pseudo ground truth across different modalities and anatomical regions. The recall curves based on different CD thresholds are shown in Fig. 2. Exemplary qualitative results are presented in

[7] https://github.com/boschresearch/direg

[8] https://docs.scipy.org/doc/scipy/reference/generated/scipy.optimize.minimize.html

[9] https://docs.scipy.org/doc/scipy/reference/generated/scipy.optimize.differential_evolution.html

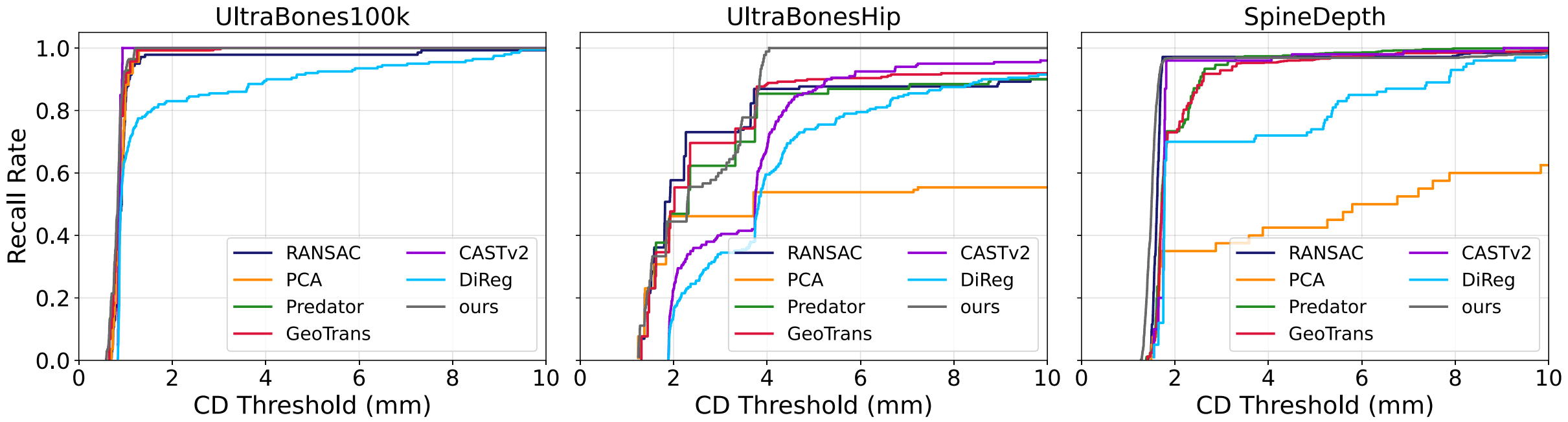

**Fig. 2.** CD recall curves for UltraBones100k (first column), UltraBonesHip (second column), and SpineDepth (last column).

**Table 2**
Quantitative results on UltraBones100k.

| Method | RRE ↓ | RTE ↓ | RR ↑ | CD ↓ | HD95 ↓ |
|---|---|---|---|---|---|
| RANSAC | 5.36 ± 20.39 | 3.67 ± 12.24 | 0.74 | 1.09 ± 1.36 | 2.72 ± 4.19 |
| PCA | 2.18 ± 10.86 | 3.07 ± 7.58 | 0.81 | 0.91 ± 0.21 | 2.20 ± 0.78 |
| Predator | 1.43 ± 1.12 | 2.36 ± 1.81 | 0.89 | 0.85 ± 0.13 | 1.78 ± 0.30 |
| GeoTrans | 1.93 ± 8.39 | 3.29 ± 12.81 | 0.79 | 0.86 ± 0.22 | 1.82 ± 0.68 |
| CASTv2 | 2.09 ± 1.30 | 1.50 ± 0.89 | 0.91 | 0.87 ± 0.03 | 2.05 ± 0.13 |
| DiReg | 17.09 ± 36.57 | 5.42 ± 10.85 | 0.57 | 1.83 ± 2.57 | 5.42 ± 9.94 |
| Ours | 1.83 ± 1.30 | 2.02 ± 1.30 | 0.89 | 0.82 ± 0.12 | 2.06 ± 0.36 |
| Pseudo ground-truth | 1.40 ± 1.02 | 1.09 ± 0.43 | 1 | 0.75 ± 0.11 | 1.63 ± 0.27 |

**Table 3**
Quantitative results on UltraBonesHip.

| Method | RRE ↓ | RTE ↓ | RR ↑ | CD ↓ | HD95 ↓ |
|---|---|---|---|---|---|
| RANSAC | 15.15 ± 44.84 | 11.18 ± 33.20 | 0.68 | 5.89 ± 17.08 | 15.41 ± 27.78 |
| PCA | 63.57 ± 73.58 | 31.36 ± 34.88 | 0.32 | 14.25 ± 15.56 | 31.98 ± 29.36 |
| Predator | 16.30 ± 34.35 | 16.61 ± 46.40 | 0.71 | 4.90 ± 8.04 | 15.69 ± 22.10 |
| GeoTrans | 8.12 ± 18.59 | 10.74 ± 27.47 | 0.77 | 4.58 ± 8.76 | 13.79 ± 17.62 |
| CASTv2 | 4.02 ± 6.17 | 2.56 ± 2.50 | 0.83 | 3.17 ± 1.34 | 11.90 ± 3.39 |
| DiReg | 18.24 ± 32.31 | 12.49 ± 25.47 | 0.42 | 8.20 ± 16.45 | 23.48 ± 36.05 |
| Ours | 1.90 ± 1.56 | 2.21 ± 0.86 | 0.88 | 2.50 ± 1.08 | 8.97 ± 4.08 |
| Pseudo ground-truth | 1.22 ± 0.81 | 1.50 ± 0.65 | 1 | 2.13 ± 0.85 | 8.49 ± 3.57 |

**Table 4**
Quantitative results on SpineDepth.

| Method | RRE ↓ | RTE ↓ | RR ↑ | CD ↓ | HD95 ↓ |
|---|---|---|---|---|---|
| RANSAC | 4.25 ± 20.68 | 3.69 ± 11.51 | 0.76 | 2.37 ± 1.42 | 5.12 ± 4.01 |
| PCA | 63.19 ± 60.96 | 13.00 ± 11.05 | 0.46 | 7.09 ± 5.33 | 18.83 ± 14.10 |
| Predator | 3.43 ± 16.39 | 1.95 ± 1.50 | 0.91 | 1.99 ± 0.84 | 5.02 ± 3.15 |
| GeoTrans | 3.77 ± 10.47 | 2.89 ± 4.75 | 0.86 | 2.12 ± 1.39 | 5.51 ± 4.02 |
| CASTv2 | 7.70 ± 31.86 | 1.95 ± 4.33 | 0.89 | 1.91 ± 0.95 | 4.33 ± 3.29 |
| DiReg | 22.82 ± 35.16 | 7.34 ± 12.91 | 0.51 | 3.28 ± 2.68 | 8.92 ± 8.67 |
| Ours | 3.78 ± 19.34 | 2.80 ± 3.75 | 0.84 | 1.78 ± 1.61 | 4.26 ± 4.58 |
| Pseudo ground-truth | 1.54 ± 1.13 | 1.78 ± 0.75 | 1 | 1.59 ± 0.07 | 3.41 ± 0.23 |

Fig. 3. 3D visualizations of registration results are provided via our repository[10]

### 5.2. Results of ablation studies

Table 5 shows the results of different coarse-to-fine configurations on the UltraBones100k dataset. The lowest RR (≈ 0.61) is obtained with $H = 1000$, $B_{\text{sparse}} = 32$, and $K = 0$, whereas the highest RR (≈ 0.91) is obtained with $H = 10000$, $B_{\text{sparse}} = 32$, and $K = 128$. In addition, overall, configurations with $K = 0$ achieve a mean RR of 0.724, a mean CD of 0.914 mm, and a mean HD95 of 2.134 mm, while configurations with $K = 128$ achieve a mean RR of 0.864, a mean CD of 0.823 mm, and a mean HD95 of 2.069 mm. Fig. 6 further visualizes the best UDF curves across iterations for different numbers of heads. For balanced performance (RR≈ 0.89) and speed (runtime ≈ 5.87 s), we choose $H = 5000$, $B_{\text{sparse}} = 32$, and $K = 64$.

Within the ablations shown in Table 6, GridVolume+NeuralReg performs best, with RREs of 4.51 ± 3.52° (UltraBones100k), 6.67 ± 9.50° (UltraBonesHip), and 4.75 ± 16.86° (SpineDepth), and RTEs of 2.44 ± 0.93 mm, 3.18 ± 5.48 mm, and 2.55 ± 2.68 mm, respectively.

Across the grid-based method, we evaluated grid resolutions $128^3$, $256^3$, and $512^3$ on the UltraBonesHip and SpineDepth datasets. Increasing the grid resolution markedly improved registration accuracy

[10] https://neuralbonereg.github.io/

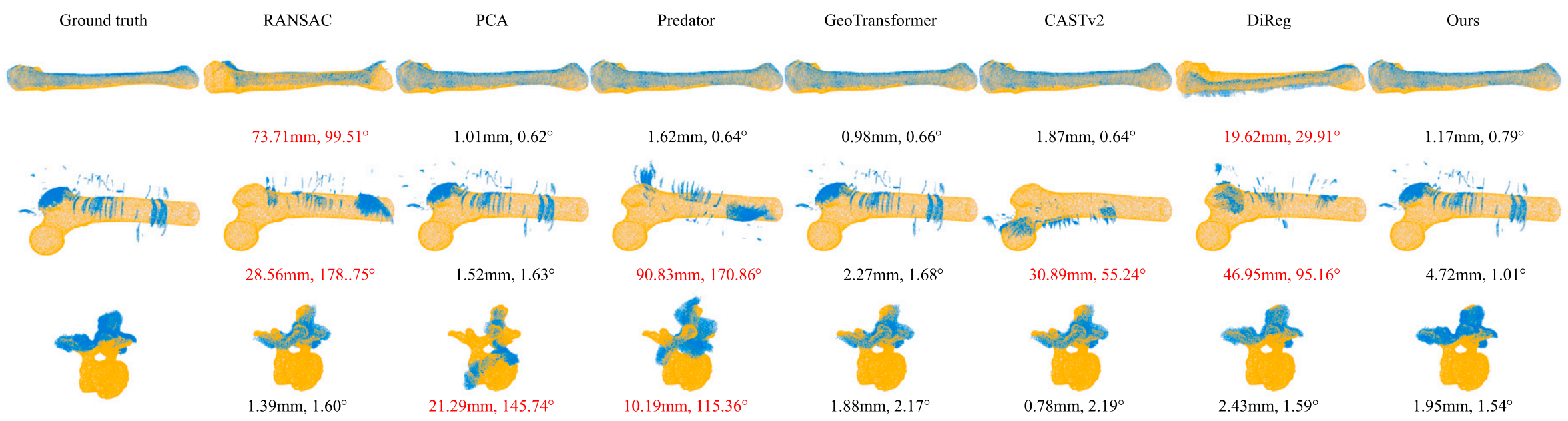

**Fig. 3.** Exemplary qualitative results. The preoperative point cloud $\mathcal{C}$ is visualized in yellow, while the intraoperative point cloud $\mathcal{U}$ is visualized in blue. The mean relative translation error (mm) and rotation error (°) are reported below each figure. Severe errors with RRE>10° or RTE>10 mm are highlighted in red..

**Table 5**
Results of different coarse-to-fine configurations on the UltraBones100k dataset. Each configuration is run 20 times per bone per specimen and optimized for 1000 iterations, yielding total FLOPs that are 1000×larger. The runtime was evaluated on a system equipped with an Intel Core i9-14900K CPU, an NVIDIA GeForce RTX 4090 GPU, and 64 GB of RAM.

| # heads $H$ | # coarse points $B_{sparse}$ | # refine heads $K$ | # params | FLOPs/iter | runtime (s) | RR | CD (mm) | HD95 (mm) |
|---|---|---|---|---|---|---|---|---|
| 1000 | 32 | 0 | 1.01M | 2.16G | 4.01 ± 0.82 | 0.609 | 0.98 ± 0.26 | 2.27 ± 0.52 |
| 1000 | 32 | 32 | 1.01M | 8.62G | 4.61 ± 0.85 | 0.842 | 0.82 ± 0.12 | 2.07 ± 0.36 |
| 1000 | 32 | 64 | 1.01M | 15.09G | 5.09 ± 0.80 | 0.854 | 0.83 ± 0.13 | 2.08 ± 0.36 |
| 1000 | 32 | 128 | 1.01M | 28.02G | 5.69 ± 0.84 | 0.817 | 0.83 ± 0.13 | 2.08 ± 0.36 |
| 1000 | 64 | 0 | 1.01M | 4.31G | 4.31 ± 0.96 | 0.795 | 0.90 ± 0.17 | 2.10 ± 0.35 |
| 1000 | 64 | 32 | 1.01M | 10.78G | 4.98 ± 1.12 | 0.838 | 0.83 ± 0.13 | 2.09 ± 0.37 |
| 1000 | 64 | 64 | 1.01M | 17.24G | 5.20 ± 1.00 | 0.829 | 0.83 ± 0.13 | 2.08 ± 0.36 |
| 1000 | 64 | 128 | 1.01M | 30.18G | 5.78 ± 0.96 | 0.829 | 0.83 ± 0.13 | 2.09 ± 0.39 |
| 1000 | 128 | 0 | 1.01M | 8.62G | 4.18 ± 0.87 | 0.727 | 0.87 ± 0.16 | 2.06 ± 0.34 |
| 1000 | 128 | 32 | 1.01M | 15.09G | 4.85 ± 1.01 | 0.824 | 0.83 ± 0.13 | 2.09 ± 0.36 |
| 1000 | 128 | 64 | 1.01M | 21.55G | 5.26 ± 0.89 | 0.823 | 0.83 ± 0.13 | 2.09 ± 0.38 |
| 1000 | 128 | 128 | 1.01M | 34.49G | 6.09 ± 1.31 | 0.808 | 0.83 ± 0.13 | 2.09 ± 0.39 |
| 5000 | 32 | 0 | 4.64M | 10.79G | 4.92 ± 1.38 | 0.628 | 0.97 ± 0.24 | 2.25 ± 0.49 |
| 5000 | 32 | 32 | 4.64M | 17.25G | 5.39 ± 1.39 | 0.878 | 0.82 ± 0.12 | 2.06 ± 0.35 |
| 5000 | 32 | 64 | 4.64M | 23.72G | 5.87 ± 1.24 | 0.886 | 0.82 ± 0.12 | 2.06 ± 0.36 |
| 5000 | 32 | 128 | 4.64M | 36.65G | 6.47 ± 1.23 | 0.866 | 0.82 ± 0.12 | 2.06 ± 0.35 |
| 5000 | 64 | 0 | 4.64M | 21.56G | 5.82 ± 1.13 | 0.815 | 0.90 ± 0.17 | 2.09 ± 0.33 |
| 5000 | 64 | 32 | 4.64M | 28.03G | 7.36 ± 1.78 | 0.881 | 0.82 ± 0.12 | 2.06 ± 0.36 |
| 5000 | 64 | 64 | 4.64M | 34.49G | 7.53 ± 1.65 | 0.882 | 0.82 ± 0.12 | 2.06 ± 0.35 |
| 5000 | 64 | 128 | 4.64M | 47.42G | 7.76 ± 1.48 | 0.894 | 0.82 ± 0.12 | 2.06 ± 0.36 |
| 5000 | 128 | 0 | 4.64M | 43.11G | 6.64 ± 0.87 | 0.756 | 0.87 ± 0.15 | 2.05 ± 0.33 |
| 5000 | 128 | 32 | 4.64M | 49.58G | 7.11 ± 1.15 | 0.881 | 0.82 ± 0.12 | 2.06 ± 0.36 |
| 5000 | 128 | 64 | 4.64M | 56.05G | 6.37 ± 0.69 | 0.885 | 0.82 ± 0.12 | 2.06 ± 0.35 |
| 5000 | 128 | 128 | 4.64M | 68.98G | 6.54 ± 0.41 | 0.884 | 0.82 ± 0.12 | 2.06 ± 0.35 |
| 10000 | 32 | 0 | 9.17M | 21.57G | 5.33 ± 0.50 | 0.640 | 0.97 ± 0.25 | 2.25 ± 0.50 |
| 10000 | 32 | 32 | 9.17M | 28.04G | 5.87 ± 0.71 | 0.886 | 0.82 ± 0.12 | 2.05 ± 0.35 |
| 10000 | 32 | 64 | 9.17M | 34.50G | 6.16 ± 0.77 | 0.895 | 0.82 ± 0.12 | 2.06 ± 0.35 |
| 10000 | 32 | 128 | 9.17M | 47.43G | 6.35 ± 0.87 | 0.906 | 0.82 ± 0.14 | 2.06 ± 0.44 |
| 10000 | 64 | 0 | 9.17M | 43.12G | 5.71 ± 0.50 | 0.802 | 0.90 ± 0.17 | 2.09 ± 0.34 |
| 10000 | 64 | 32 | 9.17M | 49.59G | 6.33 ± 1.00 | 0.894 | 0.82 ± 0.12 | 2.05 ± 0.35 |
| 10000 | 64 | 64 | 9.17M | 56.05G | 6.50 ± 0.94 | 0.881 | 0.82 ± 0.12 | 2.06 ± 0.35 |
| 10000 | 64 | 128 | 9.17M | 68.99G | 6.44 ± 0.63 | 0.876 | 0.82 ± 0.12 | 2.06 ± 0.35 |
| 10000 | 128 | 0 | 9.17M | 86.23G | 6.63 ± 0.29 | 0.744 | 0.87 ± 0.15 | 2.05 ± 0.33 |
| 10000 | 128 | 32 | 9.17M | 92.69G | 7.29 ± 0.36 | 0.895 | 0.82 ± 0.12 | 2.05 ± 0.35 |
| 10000 | 128 | 64 | 9.17M | 99.16G | 7.60 ± 0.33 | 0.887 | 0.82 ± 0.12 | 2.06 ± 0.35 |
| 10000 | 128 | 128 | 9.17M | 112.09G | 8.30 ± 0.43 | 0.898 | 0.82 ± 0.12 | 2.06 ± 0.35 |

**Table 6**
Quantitative results of the ablation studies on three datasets: UltraBones100k, UltraBonesHip, and SpineDepth. RRE: relative rotation error (degrees). RTE: relative translation error (mm).

| Method | UltraBones100k | | UltraBonesHip | | SpineDepth | |
|---|---|---|---|---|---|---|
| | RRE [°] ↓ | RTE [mm] ↓ | RRE [°] ↓ | RTE [mm] ↓ | RRE [°] ↓ | RTE [mm] ↓ |
| NeuralUDF+BFGS | 117.72 ± 77.52 | 14.06 ± 18.18 | 133.75 ± 55.54 | 44.48 ± 28.57 | 92.82 ± 63.30 | 15.46 ± 10.45 |
| NeuralUDF+DE | 17.47 ± 41.33 | 3.46 ± 4.33 | 47.98 ± 34.18 | 19.09 ± 22.01 | 67.94 ± 51.16 | 12.34 ± 9.65 |
| GridVolume+NeuralReg | 4.51 ± 3.52 | 2.44 ± 0.93 | 6.67 ± 9.50 | 3.18 ± 5.48 | 4.75 ± 16.86 | 2.55 ± 2.68 |
| GridVolume+BFGS | 115.88 ± 76.92 | 33.44 ± 58.81 | 130.80 ± 59.74 | 46.17 ± 28.29 | 98.79 ± 60.43 | 16.38 ± 9.08 |
| NeuralUDF+NeuralReg (ours) | **1.68** ± **1.18** | **1.86** ± **1.09** | **1.88** ± **1.75** | **1.89** ± **0.73** | **3.79** ± **13.74** | **2.45** ± **2.41** |

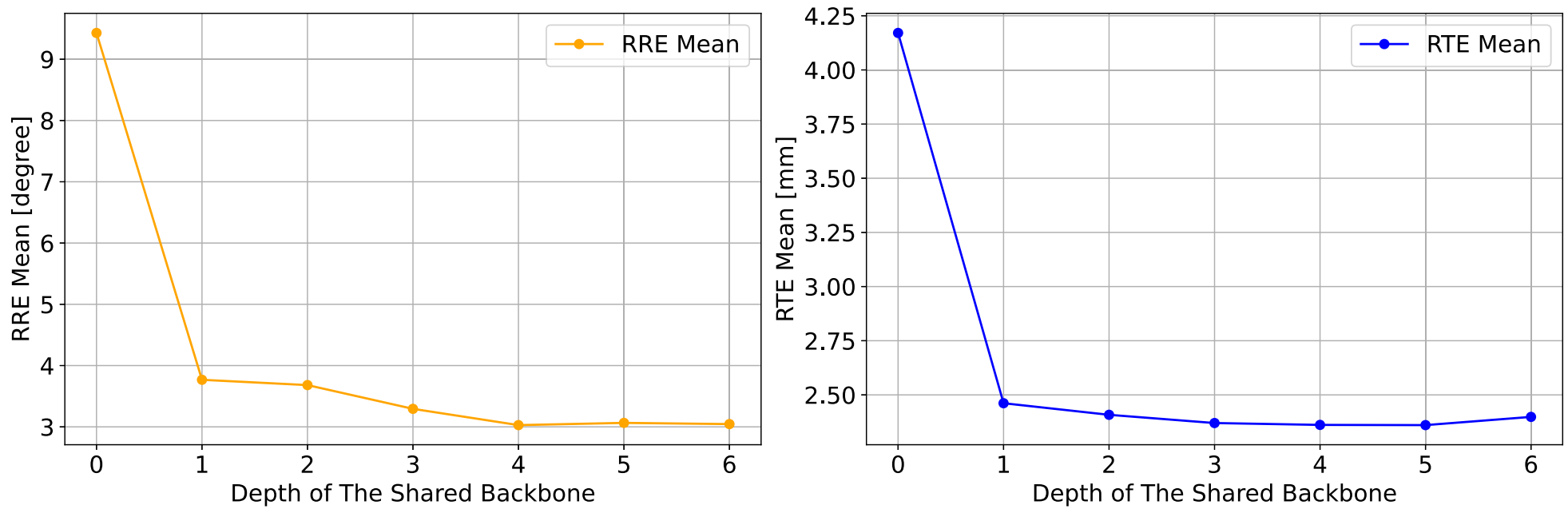

**Fig. 4.** Plots of mean RTE and RRE versus depth of the shared backbone within the NeuralReg module on the SpineDepth dataset.

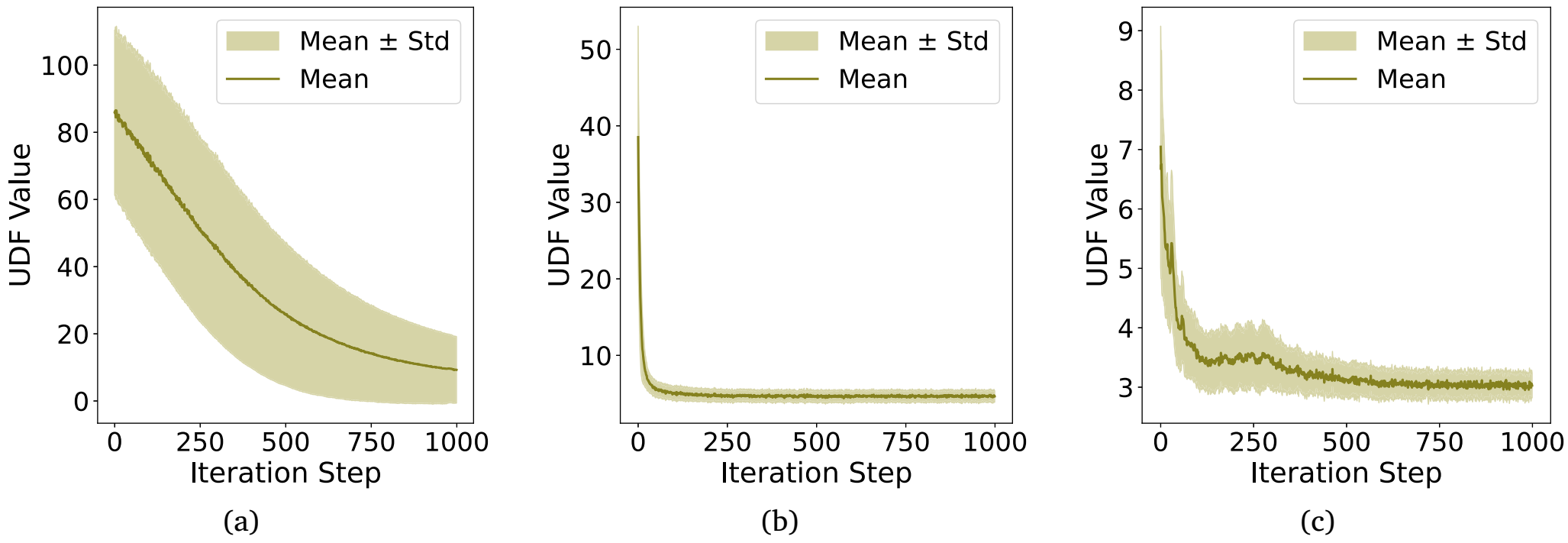

**Fig. 5.** Example UDF traces of all heads during the training process for three datasets: (a) UltraBones100k, (b) UltraBonesHip, and (c) SpineDepth.

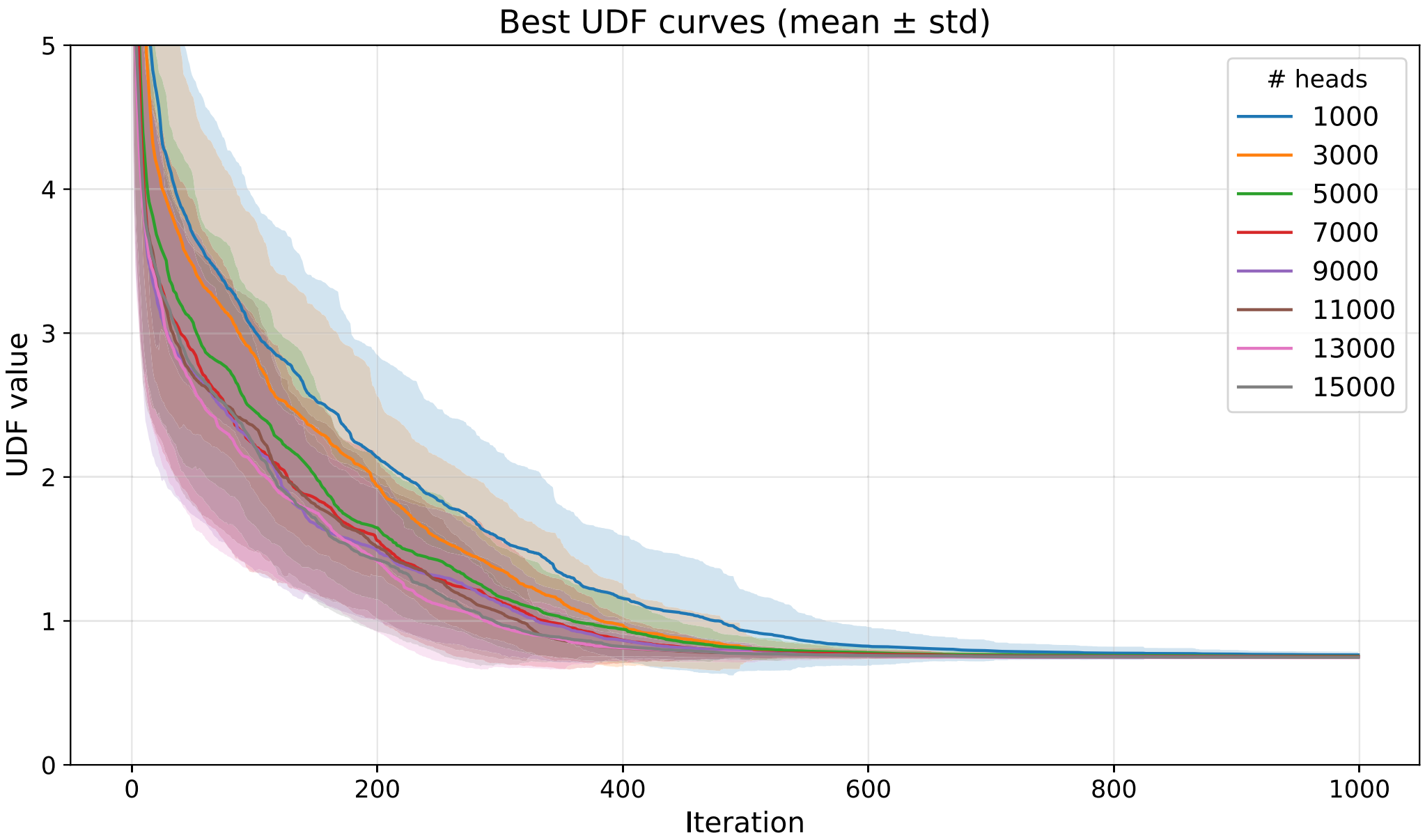

**Fig. 6.** Tracked best UDF curves across optimization iterations for different numbers of heads $H$, with fixed coarse points ($B_{\text{sparse}} = 32$) and refine heads ($K = 64$). Solid lines show the mean best UDF value across runs, and shaded bands indicate ±1 standard deviation.

for UltraBonesHip, reducing mean RRE/RTE from 31.87°/11.08 mm ($128^3$) to 11.19°/4.73 mm ($256^3$) and 6.67°/3.18 mm ($512^3$). In contrast, the improvement on SpineDepth was modest, decreasing from 5.70°/2.79 mm ($128^3$) to 4.88°/2.64 mm ($256^3$) and 4.75°/2.55 mm ($512^3$).

The effect of varying the depth of the shared backbone in the NeuralReg module is visualized in Fig. 4. Notably, when no parameters are shared between different hypothesis generation heads (depth = 0), the mean RRE and RTE are 9.43 ° and 4.17 mm, respectively, while the performance generally improves with increasing depth.

Fig. 5 depicts representative traces of UDF values across training iterations for the three datasets. Fig. 6 illustrates the impact of varying the number of heads in the NeuralReg module. To further illustrate the exploration process of heads, we visualize the SO(3) space as the axis-angle $\pi$-ball. In this ball, each rotation is represented by a 3D vector whose direction indicates the rotation axis and whose magnitude indicates the rotation angle, bounded by $\pi$. The $\pi$-ball is further discretized into $20^3$ cells for plotting 3D bins. Examples of cumulative and per-iteration $\pi$-ball coverage bins are shown in Fig. 7 and Fig. 8.

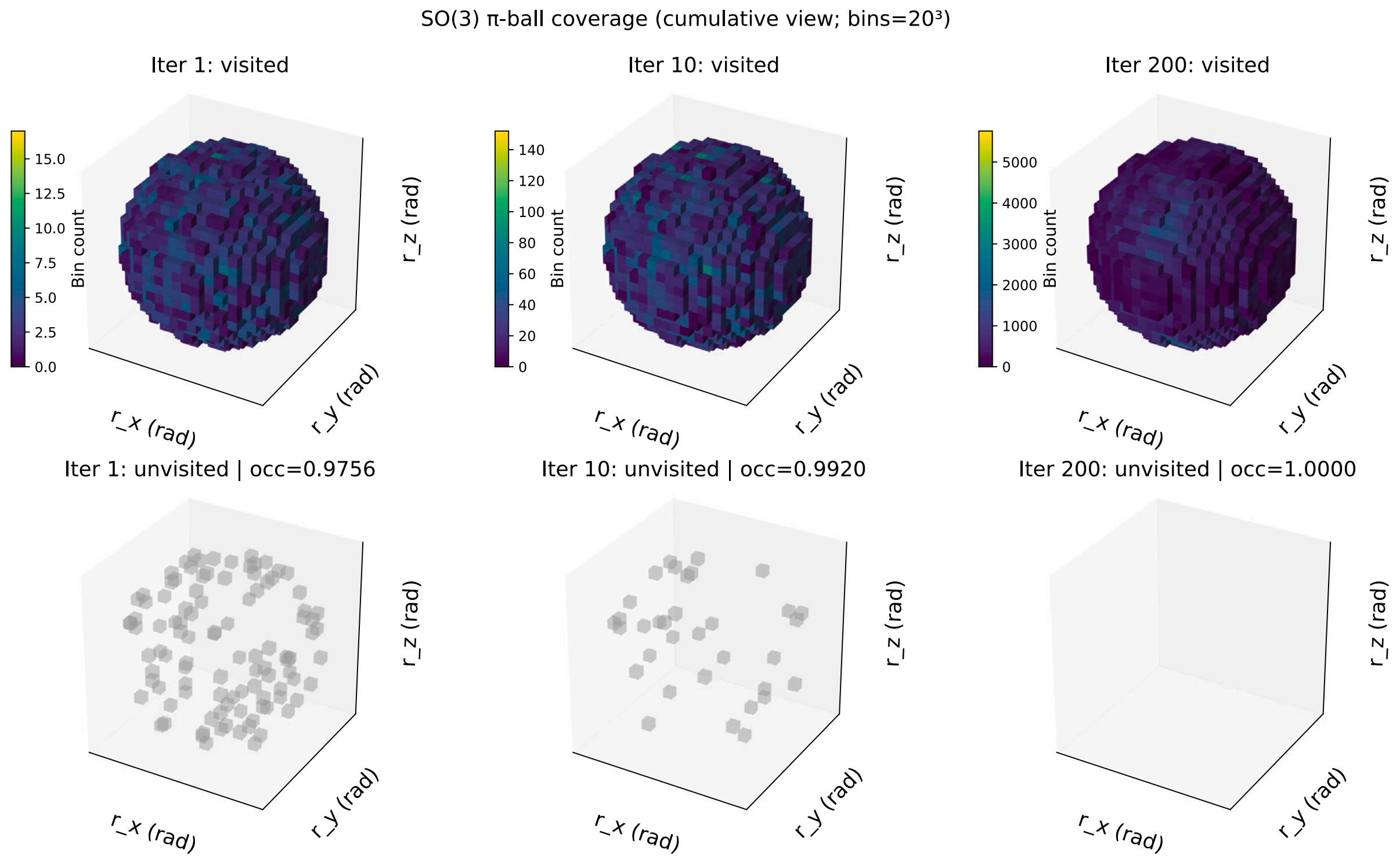

**Fig. 7.** Example cumulative SO(3) $\pi$-ball coverage at selected iterations on the UltraBones100k dataset. First row: colored voxels indicate visited bins, with color showing bin count. Second row: gray voxels indicate unvisited bins. The occupancy score, occ, is the percentage of cells visited among all valid cells inside the $\pi$-ball. Higher occ indicates broader coverage of the discretized rotation space.

**Table 7**
Results of the experiments with Gaussian noise.

| $\sigma$ | RR | CD (mean ± std) | HD95 (mean ± std) |
|---|---|---|---|
| 0.1 | 0.886 | 0.82 ± 0.12 | 2.05 ± 0.35 |
| 0.5 | 0.873 | 0.85 ± 0.11 | 2.06 ± 0.32 |
| 1.0 | 0.829 | 0.96 ± 0.10 | 2.16 ± 0.25 |
| 2.0 | 0.707 | 1.30 ± 0.10 | 2.83 ± 0.21 |
| 3.0 | 0.629 | 1.67 ± 0.12 | 3.66 ± 0.28 |
| 4.0 | 0.357 | 2.05 ± 0.14 | 4.59 ± 0.28 |
| 5.0 | 0.264 | 2.53 ± 0.69 | 5.87 ± 2.63 |

Results of the experiments with Gaussian noise are shown in Table 7. The performance degrades as $\sigma$ increases. In particular, when $\sigma$ reaches 4, the RR drops below 0.5.

## 6. Discussion

This study presents NeuralBoneReg, an instance-level, self-supervised framework for modality-agnostic bone surface registration. This method integrates implicit neural surface representations with parallel hypothesis generation to improve registration robustness in challenging CAOS scenarios. Through experiments on three multimodal datasets covering different anatomies, NeuralBoneReg demonstrated SOTA accuracy and robustness, matching or surpassing both traditional optimization-based pipelines and supervised deep learning methods. In the following, we discuss the observed performance in terms of accuracy, runtime, methodology and limitations.

**Accuracy.** The pronounced symmetry of bone anatomy leads to an optimization landscape with numerous local minima, particularly under poor initialization. SOTA registration pipelines handle these local minima differently. RANSAC+ICP attempts to mitigate them by randomized correspondence sampling, but even with 250 million iterations (≈ 226 s runtime), its results remain unstable, exhibiting high RRE/RTE variability (5.36 ± 20.39° and 3.67 ± 12.24 mm on UltraBones100k). PCA+ICP performs well when the anatomical structures exhibit distinct and stable principal axes (UltraBones100k), but degrades when PCA-derived axes become ambiguous or noisy, as in UltraBonesHip and SpineDepth. SOTA learning-based methods require large-scale paired datasets for learning robust data-driven features, including both supervised methods (requiring ground-truth registration) and unsupervised methods. Their superior performance on the UltraBones100k and SpineDepth datasets demonstrates their effectiveness. However, their performance drops significantly on the UltraBonesHip dataset due to fewer specimens and increased cross-specimen variation. For the task of bone surface registration in CAOS, the main bottleneck is data scarcity, due to the difficulty in collecting real-world paired data (Pandey et al., 2020; Hohlmann et al., 2024). To the best of our knowledge, publicly available datasets exist only for the lower extremity (UltraBones100k) and lumbar spine (SpineDepth). The lack of available datasets limits the applicability of these powerful learning-based methods to new specimens and new anatomies. In contrast, NeuralBoneReg operates self-supervised at an instance level and does not require large training datasets. Compared with the baselines, it achieves competitive or superior results across all datasets, particularly in limited-data settings.

**Runtime.** Fast execution is critical in CAOS, as prolonged intraoperative time directly impacts surgical costs and patient outcomes. Many SOTA methods perform computations during surgery and become increasingly expensive as the number of sampling iterations grows. NeuralBoneReg exploits the two-stage CAOS workflow by shifting the computationally intensive nearest-neighbor queries to the preoperative phase. Nevertheless, its intraoperative runtime (≈ 5 s) remains longer than that of SOTA methods, which complete a single forward pass within one second. However, NeuralBoneReg operates entirely self-supervised without requiring prior training datasets, offering a practical balance between computational efficiency, data independence, and generalization capability.

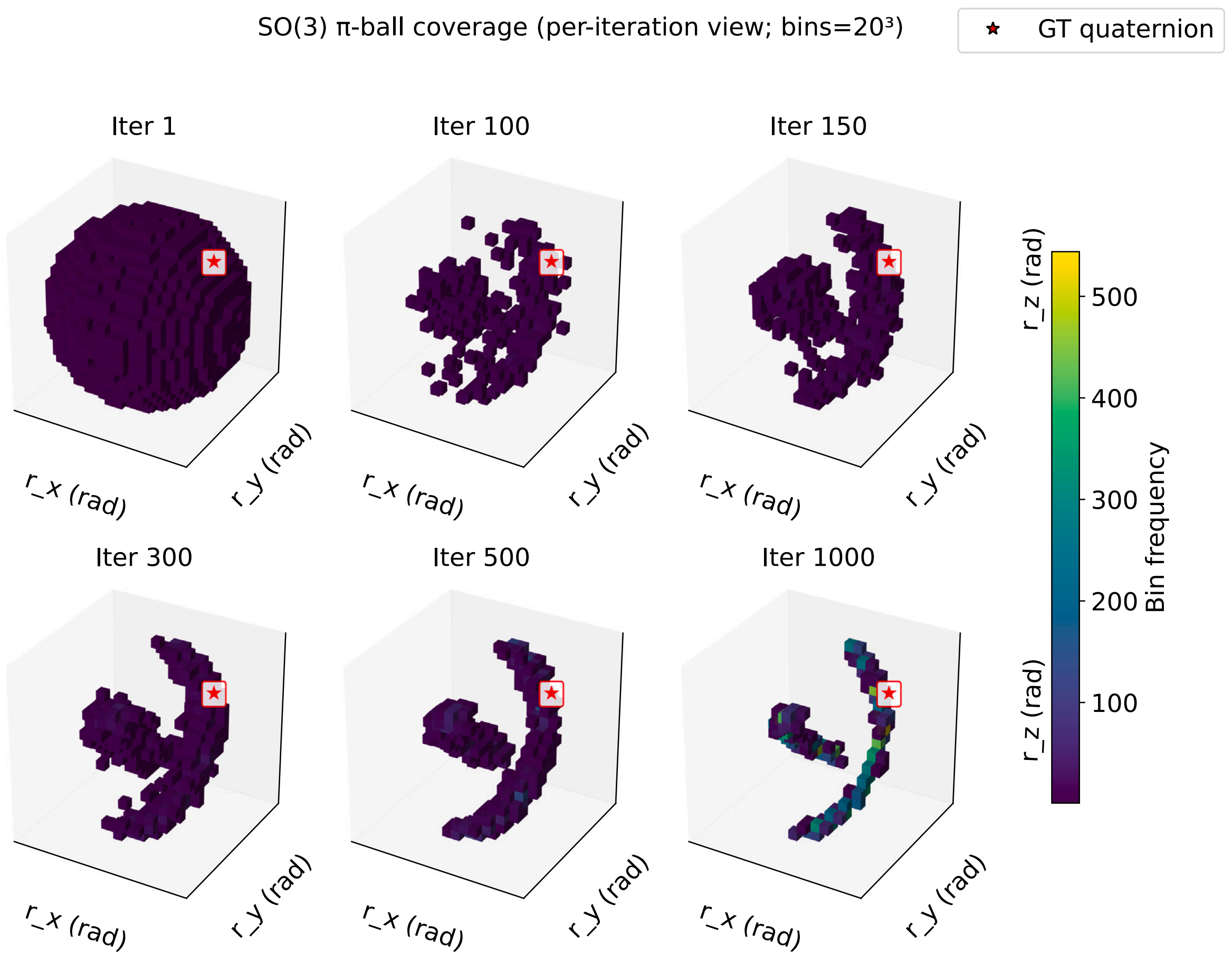

**Fig. 8.** Example per-iteration SO(3) $\pi$-ball coverage at selected iterations on the UltraBones100k dataset. Each panel shows the histogram of visited bins computed from the quaternions at that iteration only. The red star indicates the ground-truth quaternion.

**Methodology.** NeuralBoneReg combines a neural UDF with a neural optimizer for smooth and efficient registration. It differs fundamentally from random hypothesis generation and selection. During initialization, FPS improves the coverage (Fig. 7). During subsequent optimization, rather than sampling hypotheses randomly, the network learns to generate hypotheses in regions of SO(3) with low objective values under the supervision of $\mathcal{L}_{\mathrm{Reg}}$ (Eq. (10)). As shown in Fig. 5, the mean UDF across all $H$ heads converges as the number of optimization iterations increases. This trend is further supported by the converged search behavior in Fig. 8. A random hypothesis generation and selection strategy would not exhibit such convergence. In addition, Fig. 4 shows that performance degrades as the depth of the shared backbone decreases, with the largest gain from depth 0 (no sharing) to 1. These results support our hypothesis that sharing backbone parameters allows beneficial updates to propagate across heads, enabling more efficient joint learning instead of independent parallel search. The lightweight network design further enables much larger-scale parallel search than conventional methods, and the extent of parallel search directly affects robustness. In Table 6, the BFGS-based variants show large performance variability across datasets, reflecting the highly non-convex optimization landscape of the registration problem. NeuralUDF+DE mitigates this issue through population-based parallel search. Its improved performance over the BFGS-based variants confirms the advantage of exploring multiple transformation hypotheses in parallel. However, the achievable population size is constrained by computational resources: our experiments with 50 candidates exceeded the runtime of NeuralBoneReg. In contrast, the lightweight neural hypothesis-generation heads in NeuralBoneReg allow up to 20000 heads in parallel on a single PC with a 24 GB GPU. Finally, as summarized in Table 6, NeuralBoneReg achieves higher accuracy than its fixed-resolution variants, highlighting the advantage of implicit neural surface representations. Fixed-resolution distance volumes approximate distances through interpolation, which can introduce discretization artifacts and degrade gradient smoothness. As shown in Section 5.2, increasing the grid size improves performance on larger anatomies (UltraBonesHip), but yields only minor gains on smaller ones (SpineDepth), illustrating the limited scalability of fixed-resolution volumes, whose memory and runtime grow cubically with resolution. In contrast, NeuralBoneReg learns a smooth and continuous representation that supports distance queries at arbitrary resolution and enables stable, efficient gradient-based optimization through backpropagation.

**Limitations.** As an instance-specific method, NeuralBoneReg inherently requires test-time training (Sun et al., 2020), which limits its applicability to CAOS scenarios that require real-time registration. In contrast, SOTA learning-based methods do not face this limitation. In addition, NeuralBoneReg requires a complete preoperative CT-derived bone model as a reference template, limiting its applicability to more general registration problems. Table 7 shows that performance degrades with increasing input noise, highlighting limited robustness and potential room for improvement under challenging conditions. To improve robustness to noise, one potential direction is explicit denoising, including both 2D denoising of segmentation masks and 3D denoising of point clouds. Another direction is to introduce a noise-aware loss function. Currently, our method assigns equal weight to each point in Eq. (10). A more robust variant could incorporate a weight for each point based on its predicted probability of being noise. Finally, the current study evaluated bone structures only in ex-vivo datasets, in which sufficient intraoperative bone surface observations and overlap with the preoperative bone model could be obtained to support the proposed registration framework. Nevertheless, in certain specialized scenarios, such as procedures involving highly restricted surgical exposure or those in which intraoperative ultrasound acquisition is infeasible, only sparse or limited surface observations may be available, potentially reducing the applicability or performance of NeuralBoneReg. Future studies should therefore evaluate the method under such challenging partial-observation conditions and in larger in-vivo patient populations to further establish its clinical applicability.

## 7. Conclusion

In this work, we introduced NeuralBoneReg, an instance-specific self-supervised, modality-agnostic framework for bone surface registration that leverages implicit neural representations and parallel hypothesis generation. The method addresses fundamental challenges in CAOS: (i) symmetry-induced ambiguities that create unstable optimization landscapes, (ii) limited generalizability in supervised registration networks to unseen anatomies and modalities, and (iii) scalability constraints in fixed-resolution distance volumes that hinder accurate registration of large anatomical structures. By shifting computation to the preoperative stage, NeuralBoneReg enables efficient intraoperative optimization. Our findings underline the potential of instance-specific self-supervised approaches to advance CAOS by providing a more reliable foundation for surgical guidance and navigation. Future work will focus on translating our approach into clinical application.

## CRediT authorship contribution statement

**Luohong Wu:** Writing – review & editing, Writing – original draft, Validation, Software, Methodology, Formal analysis, Data curation, Conceptualization. **Matthias Seibold:** Writing – review & editing, Writing – original draft, Supervision, Project administration, Methodology, Investigation, Conceptualization. **Nicola A. Cavalcanti:** Writing – review & editing, Resources, Methodology, Data curation, Conceptualization. **Yunke Ao:** Writing – review & editing, Software, Resources, Conceptualization. **Roman Flepp:** Writing – original draft, Validation, Software. **Aidana Massalimova:** Resources, Data curation. **Lilian Calvet:** Writing – review & editing, Supervision, Methodology. **Philipp Fürnstahl:** Writing – review & editing, Supervision, Resources, Project administration, Methodology, Funding acquisition, Conceptualization.

## Declaration of Generative AI and AI-assisted technologies in the writing process

During the preparation of this work the authors used ChatGPT in order to improve the readability and language of the manuscript. After using this tool, the authors reviewed and edited the content as needed and take full responsibility for the content of the published article.

## Declaration of competing interest

The authors declare that they have no known competing financial interests or personal relationships that could have appeared to influence the work reported in this paper.

## Acknowledgments

This research has been funded by the Innosuisse Flagship project PROFICIENCY No. PFFS-21-19. This work has also been supported by the OR-X, a Swiss national research infrastructure for translational surgery, and associated funding by the University of Zurich, Switzerland and University Hospital Balgrist. This work is based on experiments performed at the Swiss Center for Musculoskeletal Imaging, SCMI, Balgrist Campus AG, Zürich.